\documentclass[11pt,letterpaper]{article}  

\usepackage[dvipsnames]{xcolor}
\usepackage{comment}
\usepackage{tikz}
\usetikzlibrary{external}
\usetikzlibrary{mindmap,trees}
\usetikzlibrary{arrows}
\usetikzlibrary{calc}
\usepackage{adjustbox}
\usepackage{subcaption}
\usepackage{epsfig}
\usepackage{epstopdf}

\usepackage{amsthm}

\usepackage{amsmath,amsfonts,amssymb,color}

\usepackage{psfrag}
\usepackage{algorithm}
\usepackage{algpseudocode}
\usepackage{array}
\usepackage{physics}
\usepackage{cite}
\date{}
\usepackage[utf8]{inputenc} 
\usepackage[T1]{fontenc}    
\usepackage{hyperref}       
\usepackage{url}            
\usepackage{booktabs}       
\usepackage{amsfonts}       
\usepackage{nicefrac}       
\usepackage{microtype}      
\usepackage{comment}
\usepackage{amsmath}
\usepackage{graphicx}
\usepackage{adjustbox}
\usepackage{textcomp}
\usepackage{xcolor}
\usepackage{tikz}
\usepackage{tikz-qtree}
\usetikzlibrary{mindmap,trees}
\usetikzlibrary{arrows}
\usetikzlibrary{calc}
\usepackage{etoolbox}

\usepackage{graphicx}
\usepackage[font={bf,it,small},labelsep=colon]{caption}
\usepackage{subcaption}
\usepackage{lipsum} 
\usepackage{caption}
\usepackage{setspace}
\usepackage{makecell}
\AtBeginEnvironment{quote}{\singlespace\vspace{-\topsep}\small}
\AtEndEnvironment{quote}{\vspace{-\topsep}\endsinglespace}
\usepackage{todonotes}
\usepackage{xspace}
\usepackage{color}
\usepackage{enumitem}
\usepackage{flushend}
\usepackage{comment}
\usetikzlibrary{tikzmark}

\usepackage[margin=1in]{geometry}                

\title{\LARGE \bf Anomaly Detection and Inter-Sensor Transfer Learning on Smart Manufacturing Datasets}

\author{Mustafa Abdallah,~Byung-Gun Joung,~Wo Jae Lee\thanks{The work conducted while the author was with Purdue University.},\\~Charilaos Mousoulis,~John W. Sutherland, and~Saurabh Bagchi
\thanks{Mustafa Abdallah, Byung-Gun Joung, Charilaos Mousoulis, John W. Sutherland, and Saurabh Bagchi are with Purdue University, West Lafayette, Indiana, USA, 47907. Email: {\tt \{abdalla0,bjoung,cmousoul,jwsuther,sbagchi\}@purdue.edu}. Wo Jae Lee is with Amazon. Email: {\tt{wojaelee@gmail.com}}. This work is under review with NeurIPS 2022 Datasets and Benchmarks track.}%
}

\begin{document}
\maketitle

\vspace{-1mm}
\begin{abstract}
Smart manufacturing systems are being deployed at a growing rate because of their ability to interpret a wide variety of sensed information and act on the knowledge gleaned from system observations. In many cases, the principal goal of the smart manufacturing system is to rapidly detect (or anticipate) failures to reduce operational cost and eliminate downtime. This often boils down to detecting anomalies within the sensor date acquired from the system. 
The smart manufacturing application domain poses certain salient technical challenges. In particular, there are often multiple types of sensors with varying capabilities and costs. The sensor data characteristics change with the operating point of the environment or machines, such as, the RPM of the motor. The  anomaly detection process therefore has to be calibrated near an operating point. In this paper, we analyze four datasets from sensors deployed from manufacturing testbeds. We evaluate the performance of several traditional and ML-based forecasting models for predicting the time series of sensor data. Then, considering the sparse data from one kind of sensor, we perform transfer learning from a high data rate sensor to perform defect type classification. Taken together, we show that predictive failure classification can be achieved, thus paving the way for predictive maintenance.
\end{abstract}

\section{Introduction}

Smart manufacturing application domain poses certain salient technical challenges for the use of ML-based models for anomaly detection. 
First, in smart manufacturing domain, there are multiple types of sensors concurrently generating data about the same (or overlapping) events. These sensors are of varying capabilities and costs. Second, the sensor data characteristics change with the operating point of the machines, such as, the RPM of the motor. The inferencing and the anomaly detection processes therefore have to be calibrated for the operating point. Thus, we need case studies of anomaly detection deployments on such systems --- the need for such deployments and resultant analyses have been made 
for smart manufacturing systems~\cite{thomas2018minerva,scime2018anomaly} (see also the survey  \cite{wang2018deep} on the usage and challenges of deep learning in smart manufacturing systems).  Most of the existing work has relied on classical models for anomaly detection and failure detection in such systems \cite{7474197,shahzad2016energy,mitchell2014survey,chatterjee2020context}. While there is a rich literature on anomaly detection in many IoT-based systems~\cite{chandola2009anomaly,sabahi2008intrusion}, there are few existing works that document the use of ML models for anomaly detection 
in smart manufacturing systems~\cite{s20071813} (see~\cite{lopez2017categorization} for a survey). In particular, most of the existing work focused on categorizing anomalies in the semi-conductor industry \cite{susto2017anomaly}, windmill monitoring \cite{leahy2016diagnosing}, and laser-based manufacturing~\cite{FRANCIS201910}.

There is also important economic impetus for this kind of deployment and analysis. In a smart manufacturing system, various sensors (e.g., vibration, ultrasonic, pressure sensors) are applied for process control, automation, production plans, and equipment maintenance. For example, in equipment maintenance, the condition of operating equipment is continuously monitored using proxy measures (e.g., vibration and sound) to prevent unplanned downtime and to save maintenance costs \cite{lee2019mfg1}. Thus, the data from these sensors can be analyzed in a streaming, real-time manner to fill a critical role in predictive maintenance tasks, through the anomaly detection process~\cite{garcia2006simap,kroll2014system,de2018anomaly}. 
Thus, we propose our anomaly detection technique for 
smart manufacturing systems~\cite{kusiak2018smart}. 
Two notable exceptions to the lack of prior work in this domain are the recent works \cite{lee2019mfg2,ALFEO2020272}. 
In \cite{lee2019mfg2}, the authors proposed a kernel principal component analysis (KPCA)-based anomaly detection system to detect a cutting tool failure in a machining process. The work~\cite{ALFEO2020272} provided a deep-learning based anomaly detection approach. However, they did not address the domain-specific challenges introduced above, did not propose any learning transfer across different manufacturing sensors as we propose here, and did not benchmark the performance of diverse forecasting models for the anomaly detection task.

\noindent {\bf Our Contribution}: \\
In this paper, 
we study the maintenance problem of smart  manufacturing systems by detecting failures and anomalies that would have an impact on the reliability and security of these systems.
In such systems, the data are collected from different sensors via intermediate data collection points and finally aggregated to a server to further store, process, and perform useful data-analytics on the sensor readings \cite{marjani2017big,he2017multitier}. We propose a \emph{temporal anomaly detection} model, in which the temporal relationships between the readings of the sensors are captured via a time-series prediction model. Specifically, we consider two classes of time-series prediction models which are classical forecasting models (including \textit{\bf Autoregressive Integrated Moving Average model (ARIMA)}~\cite{contreras2003arima}, Seasonal Naive~\cite{montero2020fforma}, and Random Forest~\cite{liaw2002classification})  and new ML-based models (including \textit{\bf Long Short-Term memory (LSTM)}~\cite{818041}, \textit{\bf Auto Encoder}~\cite{chollet2016building}, and \textit{\bf DeepAR}~\cite{salinas2020deepar}). These models are used to predict the expected future samples in certain time-frame given the near history of the readings. We first test our models on real data collected from deployed manufacturing sensors to detect anomalous data readings. We then analyze the performance of our models, and compare the algorithms of these time-series predictors for different testbeds. We observe that the best forecasting model is dataset-dependent with ML-based models giving better performance in the anomaly detection task.


 Another problem in this domain is the prediction from models using sparse data, which is often the case because of limitations of the sensors or the cost of collecting data. 
 One mitigating factor is that plentiful data may exist in a slightly different context, such as, from a different kind of sensor on the same equipment or the equipment being operated under a somewhat different operating condition in a different facility (such as a different RPM). Thus, the interesting research question in this context is: \textit{can we use a model trained on data from one kind of sensor (such as, a piezoelectric sensor, which has a high sampling frequency) to perform anomaly detection on data from a different kind of sensor (such as, a MEMS sensor, which has a low sampling frequency but is much cheaper)}. In this regard, we propose a transfer-learning model that transfers learning across different instances of manufacturing vibration sensors. This transfer-learning model is based on sharing weights and feature transformations from a deep neural network (DNN) trained with data from the sensor that has a high sampling frequency. These features and weights are used in the classification problem of another sensor data.\footnote{By classification problem here, we mean doing both re-training on the new sensor using the shared neural weights and the feature representation and then doing the defect type classification.} the one with lower sampling frequency. We show that the transfer-learning idea gives a relative improvement of 11.6\% in the accuracy of classifying the defect type over the regular DNN model. We built variants of DNN models for the defect classification task, i.e., using a single RPM data for training and for testing across the entire operating environment, and using aggregations of data across multiple RPMs for training with interpolation within RPMs.
One may wonder why we need to use sensors with much lower sampling rate; the reason is the significant price difference between the MEMS sensor and piezoelectric sensor. The former has much lower resolution (and also cost \cite{VibrationReport,albarbar2008sensor}---\$8 versus \$1,305). 
Therefore, the goal is to build a predictive maintenance model from the piezoelectric sensor and use it for the MEMS sensor.

In this paper, we test the following hypotheses related to anomaly detection in smart manufacturing. 

\textbf{Hypothesis 1:} Deep learning-based anomaly detection technique is effective for smart manufacturing.
\textbf{Hypothesis 2:} Learning process for classifying failures is transferable across different sensor types.

Based on our analysis with real data, we have the following contributions: 
\vspace{-6pt}
\begin{enumerate}[noitemsep,topsep=2pt,parsep=0pt,partopsep=2pt,leftmargin=*]
\item \textbf{Anomaly Detection:} We build two classes for time series prediction models for temporal anomaly detection on real sensor data in smart manufacturing system for detecting anomaly readings collected from the deployed sensors. 
We test our models for temporal anomaly detection through four real-world datasets collected from manufacturing sensors (e.g., vibration data). We observe that the ML-based models outperform the classical models in the anomaly detection task. 
 
\item \textbf{Defect Type Classification:} We detect the level of defect (i.e., normal operation, near-failure, failure) for each RPM data using deep learning (i.e., deep neural network multi-class classifier) and we transfer the learning across different instances of manufacturing sensors. 
We analyze the different parameters that affect the performance of prediction and classification models, such as the number of epochs, network size, prediction model, failure level, and sensor type. 

\item \textbf{RPM Selection and Aggregation:} We show that training at some specific RPMs, for testing under a variety of operating conditions gives better accuracy of defect prediction. The takeaway is that careful selection of training data by aggregating multiple of predictive RPM values is beneficial. 

\item \textbf{Benchmark Data:} We release our database
corpus (4 datasets) and codes for the community to access it for
anomaly detection and defect type classification and to build on it with new datasets and models.\footnote{URL for our database and codes is: \\ \url{https://drive.google.com/drive/u/2/folders/1QX3chnSTKO3PsEhi5kBdf9WwMBmOriJ8}} We are unveiling real failures of a pharmaceutical packaging manufacturer company.

\end{enumerate}

\section{Proposed Models} \label{sec:model}

We now describe our proposed algorithms for the anomaly detection and defect type classification.

\subsection{Temporal Anomaly Detection}
Here, we describe our proposed algorithm for detecting anomalies from the sensor readings. First, we build time-series predictors, using different time-series predictor variants in our algorithm. We compare several state-of-the-art time-series forecasting models for our anomaly detection task on our manufacturing testbeds. They can be classified into the following two classes:
\begin{itemize}[leftmargin=*]
\vspace{-1mm}
    \item \textbf{Classical forecasting models:} In this category, we included Autoregressive Integrated Moving Average model (ARIMA)~\cite{contreras2003arima}, Seasonal Naive~\cite{montero2020fforma} (in which each forecast equals the last observed value from the same season), Random Forest (RF)~\cite{liaw2002classification} (which is a tree ensemble that combines the predictions made by many decision trees into a single model), and Auto-regression~\cite{lewis1985prediction}.  
    \item \textbf{ML-based forecasting models:}
     We selected six popular time series forecasting models, including Recurrent Neural Network (RNN)~\cite{tokgoz2018rnn}, LSTM~\cite{818041} (which is a better version than RNN and has been used in different applications~\cite{graves2005framewise,abdallah2019athena}), Deep Neural Netowrk (DNN)~\cite{sen2019think}, AutoEncoder~\cite{chollet2016building}, and the 
recent works DeepAR~\cite{salinas2020deepar}, DeepFactors~\cite{wang2019deep}.
\end{itemize}

For each model, we generated multiple variants by varying the values of hyperparameters and we chose the model variant with the best performance for each dataset. We describe the hyper-parameters and the libraries used for all forecasting models in Appendix E (in the supplementary material).

\noindent \textbf{Anomaly Detection Rule}: 
After using any of the above proposed time-series predictors, for each sample under test, we would have two values: the actual value (measured by the sensor) and the predicted value (predicted by our model). To flag an anomaly, we consider that $\frac{\text{predicted value} - \text{actual value}}{\text{predicted value}} > \lambda$. In other words, the relative error between the actual value and the predicted value is more than $\lambda$.\footnote{In our work, we also used classifier-based model for  anomaly detection of test samples (See Appendix F.2).} 

\subsection{Transfer Learning across Sensor Types}\label{proposed_model_transfer_learning}
\begin{figure*}[ht]
\vspace{-3mm}
\centering
  \includegraphics[width=0.85\textwidth]{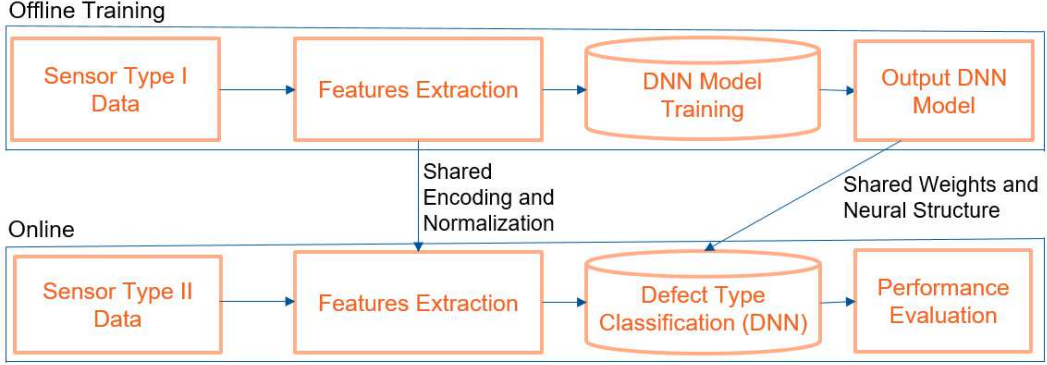}
  \caption{The proposed learning-transfer model  has two modes: offline DNN sub-model training and online-mode for classifying the sensor under test after sharing knowledge (i.e., DNN's weights and features).} 
  \label{fig:learning-transfer model}
  \vspace{-6pt}
\end{figure*}

We show our proposed model in Figure~\ref{fig:learning-transfer model} which has two modes: In offline training, the sensor with large amount of data (let us call it sensor type I) has its data entered to the feature extraction module that performs encoding and normalization of the input signals into numerical features. Second, a deep neural network (DNN) model is trained and tuned using these features and labels of the data 
(normal, near-failure or failure). We use the DNN as a multi-class classifier due to its discriminative power that is leveraged in different classification applications %
\cite{yuan2016deepgene,ahmed2017unsupervised,al2018computer,elaraby2016deep}. Moreover, DNN is useful for both tasks of learning the level of defect for the same sensor type and for transfer learning across the different sensor types that we consider here.
In online mode, any new sensor data under test (here, sensor type II) would have the same feature extraction process where the saved feature encoders are shared. Then, the classifier (after retraining) predicts the defect type (one of the three states mentioned earlier) given the trained model, and giving as output the probability of each class. 
 
 It is worth noting that sensor types I and II should be measuring the same physical quantity but can be from different manufacturers and with different characteristics. For instance, in our smart manufacturing domain, sensor type I is a piezoelectric sensor (of high cost  but with high sampling resolution) while type II is a MEMS sensor (of lower cost but with lower sampling resolution). We propose the transfer learning for predictive maintenance i.e., predicting the level of defect of the MEMS sensor and whether it is in normal operation, near-failure (and needs maintenance), or failure (and needs replacement). We emphasize that although the two sensor types we consider for that task in our work generate different data distribution and have different sampling frequency, our transfer learning is efficient (see our evaluation in Section~\ref{sec:learning_tranfer_sensors_results}).

Having introduced the background and the high-level proposed models.  We next detail the anomaly detection and the transfer learning tasks on our manufacturing testbeds, respectively.
\section{Anomaly Detection with Manufacturing Sensors}\label{sec:anomaly_detect_manuf}

Anomalous data generally needs to be separated from machine failure as abnormal patterns of data do not necessarily imply machine or process failure \cite{WANG2018144}. We perform anomaly detection using vibration and process data to identify anomalous events and then attempt to label/link these events with machine failure information. This way, we aim to identify abnormal data and correlate the abnormal data to machine failure coming from manufacturing sensors. The manufacturing of discrete products typically involves the use of equipment termed machine tools. Examples of machine tools include lathes, milling machines, grinders, drill presses, molding machines, and forging presses. Almost always, these specialized pieces of equipment are reliant on electric motors that power gearing systems, pumps, actuators, etc. The health of a machine is often directly related to the health of the motors being used to drive the process. Given this dependence, health studies of manufacturing equipment may work directly with equipment in a production environment or in a more controlled environment on a "motor testbed." 
To achieve such goal, we build time-series models to predict (and detect) anomalies in the sensors. We first detail our datasets.
\vspace{-1mm}
\subsection{Deployment Details and Datasets Explanation}\label{sec:dataset_description}

\begin{figure*}[t] 
\begin{minipage}[t]{1.0\textwidth}
\begin{minipage}[t]{.32\textwidth}
\centering
\includegraphics[width=0.5\linewidth]{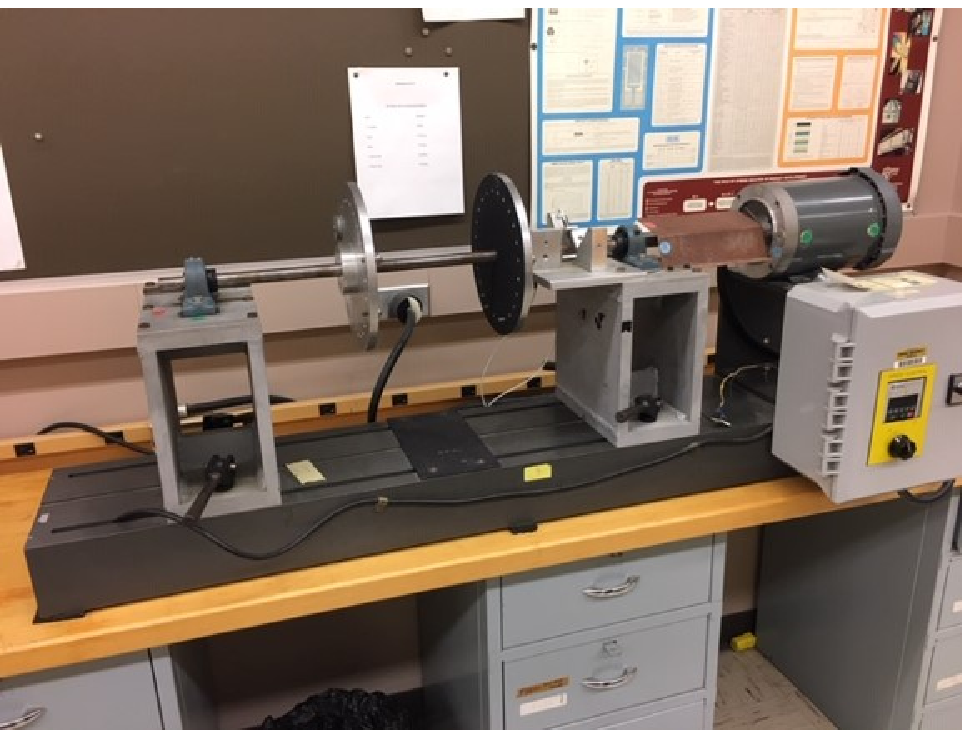}
  \caption{Motor Testbed.}
  \label{fig:testbed1} 
\end{minipage}\hfill
\begin{minipage}[t]{.32\textwidth}
\centering
  \includegraphics[width=0.6\linewidth]{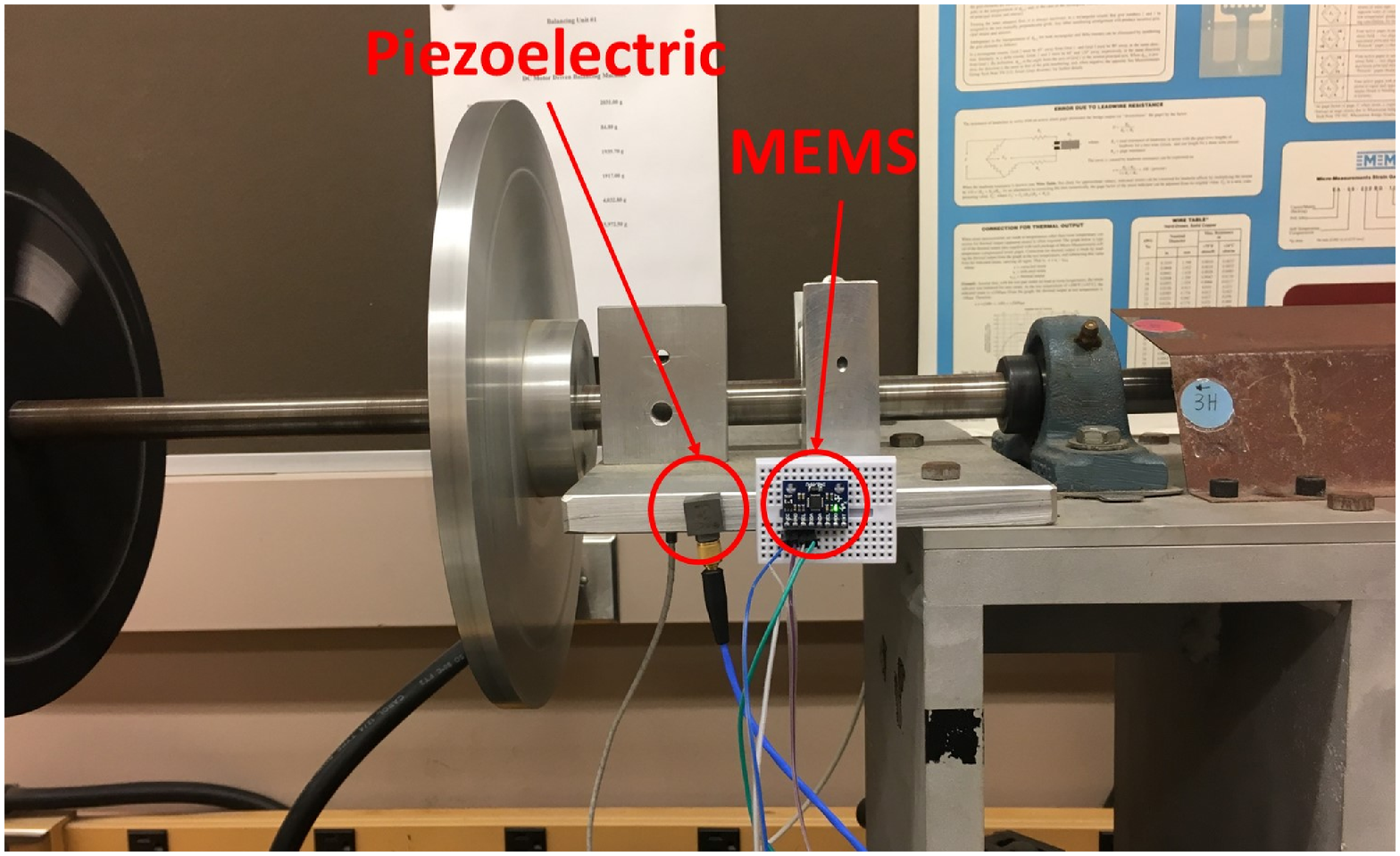}
  \caption{Piezoelectric and MEMS  sensors mounted on motor testbed.}
  \label{fig:testbed2}
\end{minipage}\hfill
\begin{minipage}[t]{.32\textwidth}
\centering
\includegraphics[width=0.5\linewidth]{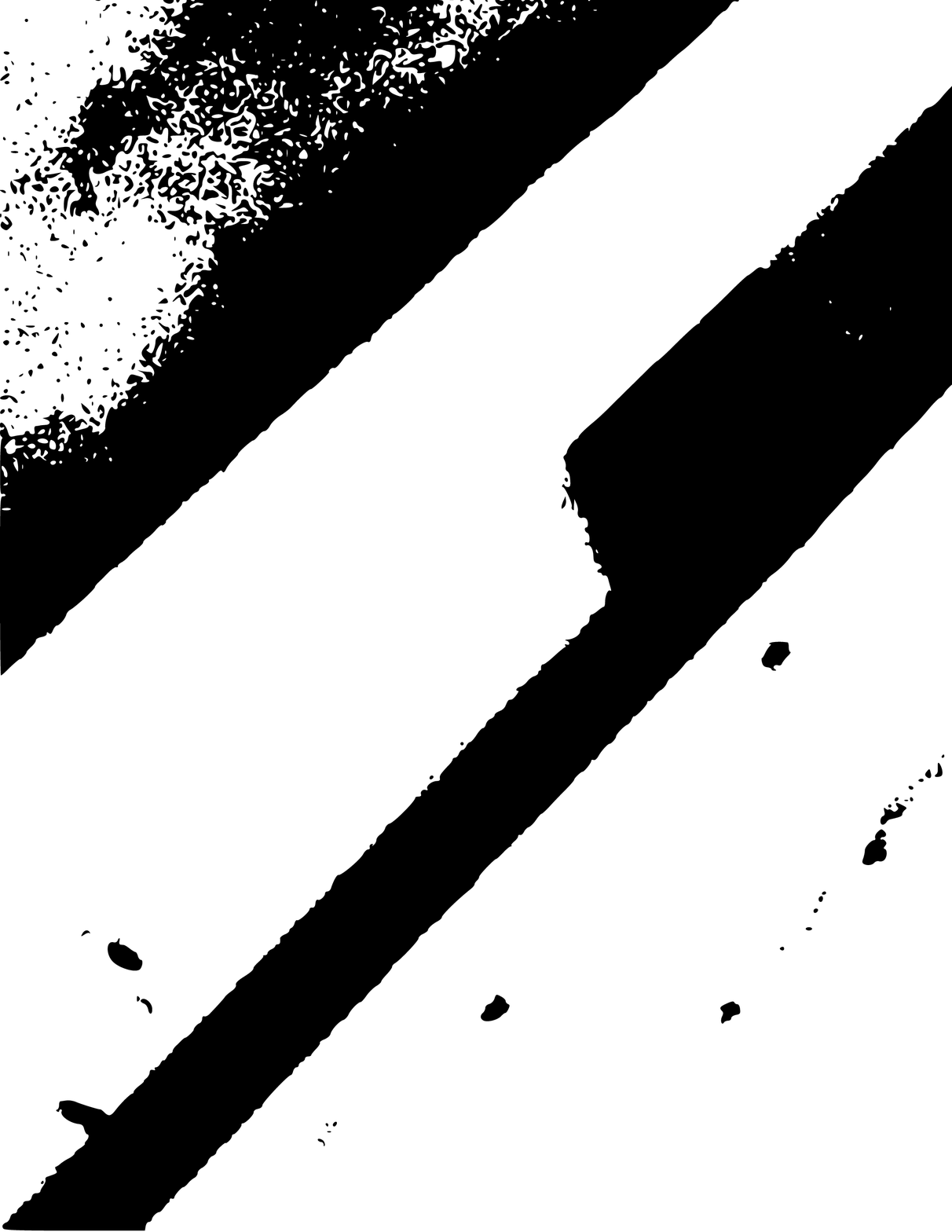}
  \caption{Balancing disk to make different levels of imbalance.} 
  \label{fig:testbed3}
\end{minipage}
\end{minipage}
\vspace{-0.28in}
\end{figure*}

\textbf{(1) Piezoelectric and MEMS datasets:} To build these datasets, an experiment was conducted in the motor testbed (shown in Figure \ref{fig:testbed1}) to collect machine condition data (i.e., acceleration) for different health conditions. During the experiment, the acceleration signals were collected from both piezoelectric and MEMS sensors (Figure \ref{fig:testbed2}) at the same time with the sampling rate of 3.2 kHz and 10 Hz, respectively, for X, Y, and Z axes. Different levels of machine health condition can be induced by mounting a mass on the balancing disk (shown in Figure \ref{fig:testbed3}), thus different levels of mechanical imbalance are used to trigger failures. Failure condition can be classified as one of three possible states - normal, near-failure, and failure.
Acceleration data were collected at the ten rotational speeds (100, 200, 300, 320, 340, 360, 380, 400, 500, and 600 RPM) for each condition. While the motor is running, 50 samples were collected at 10 second interval, for each of the ten rotational speeds. We use this same data for defect-type classification and learning transfer tasks (Section~\ref{sec:learning_tranfer_sensors_results}). 

(2) \textbf{Process and Pharmaceutical Packaging datasets:}
In the production of injection molded plastic components, molten material is injected into a die. To increase the production rate (i.e., speed up the process), a coolant is circulated through a piping system embedded within the die to remove heat from the system.  This accelerates the rate at which the die and plastic components cool and solidify, and reduces the cycle time.  Of course, the coolant within this system must then have heat removed from it; this is often achieved with the aid of a chiller. Discussions with our company partner (makes plastic containers for the pharmaceutical industry) indicated that there might be concerns with the vibration of the chiller. Data were collected on the chiller vibration. We were also able to collect process related data that can potentially indicate the condition of machine operation. Such process data is being collected as part of the company's standard statistical process control (SPC) activities; 49,706 samples of process data were collected for the period from Aug. 2021 - May 2022. One type of process data collected was the internal temperature of the chiller for the injection molding machines. In this paper, the chiller temperature was used for anomaly detection task. The chiller in the pharmaceutical process is designed to maintain the temperature of the cooling water used in the manufacturing process to around 53 Fahrenheit degree. When the chiller operation is down, the temperature of the process water varies with the ambient temperature. The sampling rate of the process data is 1 data point per 5 minutes when the SPC system is on service. As the pharmaceutical company operates non-stop from Sunday 11 pm to Friday 7 pm and shuts down from Friday 7 pm to Sunday 11 pm.  When the chiller is failed, supply temperature can vary and goes up to 65 degrees.

\textbf{Experimental Setup:} The goal is to measure the performance of our time-series regression model to detect anomalies for the vibration sensors. We show the performance of our models in terms of the accuracy of detecting anomalies (measured by precision, recall, and F-1 score). We also use the root mean square error (RMSE) for evaluating the performance of different forecasting models on the four datasets. The goal of these time-series regression models is to extract the anomaly measures that are typically far from the predicted value of the regression model.
 For each proposed model, the training size was 66\% of the total collected data while the testing size was 34\%. 
 We also varied the proportion of data used for training as a parameter and tested the performance of our model to check the least amount of data needed (which which was 30\% of the data in our experiments) for the time-series regression model to predict acceptable values (within 10\% error from the actual values).
We trained the ten predictive models on specific RPM and tested on same RPM. The data contains different levels of defects (i.e., different labels for indicating normal operation, near-failure, and failure). These labels would be used in next section. In time-series prediction models, all data that have different levels of defects were tested. Specifically, the data was divided between training and testing equally. We stopped after 5 epochs as the total loss on training samples saturates.

\textbf{Computing Resources:} We performed anomaly detection experiments on an Intel i7 @2.60 GHz, 16GB RAM, 8-core workstation. The transfer learning experiments were performed on Dell Precision T3500 Workstation, with 8 CPU cores, each running at 3.2GHZ, 12GB RAM, and Ubuntu 16.04 OS. 

\subsection{Results and Insights} 

\textbf{Performance:}
We first do bench-marking of the ten time-series forecasting models for each of the four datasets (described above in Section~\ref{sec:dataset_description}). Table~\ref{table:results_uv} shows such comparison in terms of the RMSE. We first observe that each dataset has a different best model (e.g., LSTM gave the best performance for Piezoelectric dataset while AutoEncoder was the best for Process data). Second, most of the ML-based forecasting models perform better than the traditional models. This is due to the fact that the deployments generate enough data for accurate training and due to the complex dependencies among the features of the datasets. Third, the linear models such as ARIMA and Auto-Regression were worse due to the non-linear nature of sensors' data. We second compare the anomaly detection performance of our approach under the different forecasting models (represented by the typical metrics: Precision, Recall, and F-1 Score~\cite{campos2016evaluation}). Table~\ref{table:results_precision_recall} shows the average performance for each metric across our four datasets. We observe that Random Forest and AutoEncoder give the first and second best anomaly detection performances, respectively (i..e., highest precision and recall). Also, Seasonal Naive and Auto Regression gave the worst performance.

\begin{table*}[ht!]
	\centering
	\caption{Results for  forecasting (RMSE; the lower the better) for every testbed. For each forecasting model, we choose the model with the best performance from all its model variants. We observe that the best forecasting model is task-dependent (i.e., the best model is varying depending on each dataset type).}
	\label{table:results_uv}
	\small
	\footnotesize
	\vspace{-3mm}
	\resizebox{0.99\textwidth}{!}{
	\begin{tabular}{l c c c c c c c c c c c c}
		\toprule
		 Dataset & Seasonal Naive &
		DeepAR & 
		Deep Factors &
		Random Forest & 
		AutoEncoder & 
		Auto-Regression  & 
	    ARIMA &
        LSTM &
        RNN &
        DNN  \\ 
	    \hline
         \textbf{Piezoelectric} & $0.0834$ & $0.0849$ & $0.0813$ & $0.0554$ & $0.0931$ & $0.3804$ &  $0.0954$ & $\textbf{0.0340}$ & $0.0352$ & $0.0390$ \\

    
         \textbf{MEMS} & $0.1442$ & $\textbf{0.1346}$ & $0.2510$ & $0.2295$ & $0.2431$ & $0.3865$ &  $0.2569$ & $0.1450$ & $0.1501$ & $0.1554$ \\

	    
         \textbf{Process Data} & $0.8943$ & $0.8262$ & $6.6468$ & $0.5357$ & $\textbf{0.0560}$ & $1.1645$ &  $1.8840$ & $0.6001$ & $0.5811$ & $0.769$ \\

	   
         \textbf{Pharmac. Packaging} & $0.5673$ & $0.3654$ & $0.3628$ & $0.1597$ & $\textbf{0.1510}$ & $0.3962$ &  $1.3069$ & $0.7031$ & $0.7612$ & $1.5820$ \\
	
		\bottomrule
	\end{tabular}
	}
	\vspace{-4mm}
\end{table*}

\begin{table*}[ht!]
	\centering
	\caption{Anomaly detection Results (Precision, Recall, and F-1 Score; the higher the better) for each forecasting model. Random Forest and AutoEncoder give the best anomaly detection performances.}
	\label{table:results_precision_recall}
	\small
	\footnotesize
	\vspace{-3mm}
	\resizebox{0.99\textwidth}{!}{
	\begin{tabular}{l c c c c c c c c c c c}
		\toprule
		 Metric & Seasonal Naive &
		DeepAR & 
		Deep Factors &
		Random Forest & 
		AutoEncoder & 
		Auto-Regression & ARIMA &
        LSTM &
        RNN &
        DNN \\ 
	    \hline
         \textbf{Precision} & $0.4285$ & $0.5609$ & $0.5409$ & $\textbf{0.8333}$ & $0.7833$ & $0.5116$ & $0.7419$ & $0.5135$ & $0.5961$ & $0.6305$ \\
	     
	     \textbf{Recall} & $0.4502$ & $0.6571$ & $0.6071$ & $\textbf{0.7813}$ & $0.7705$ & $0.5945$ & $0.6216$ & $0.5938$ & $0.6818$ & $0.6744$ \\
	     
	     \textbf{F-1 Score} & $0.4391$ & $0.6052$ & $0.5721$ & $\textbf{0.8064}$ & $0.7769$ & $0.5499$ & $0.6764$ & $0.5507$ & $0.6361$ & $ 0.6517$ \\

		\bottomrule
	\end{tabular}
	}
	\vspace{-4mm}
\end{table*}

\section{Transfer Learning across Vibration Sensors} \label{sec:learning_tranfer_sensors_results}

In this section, we use our transfer-learning proposed model to detect the level of defect of the readings from the manufacturing sensors. In this context, we evaluate the performance of the model on two real datasets from our manufacturing sensors which are piezoelectric and MEMS vibration sensors. In other words, we perform data analytics on the data from the vibration sensors and infer one of three operational states (mentioned in Section~\ref{sec:anomaly_detect_manuf}) for the motor. We show the performance of our model in terms of the accuracy of detecting defect level as measured by the classification accuracy of the deep-learning prediction model on the test dataset which is the defined as the number of correctly classified samples to the total number of samples. We study different parameters and setups that affect the performance. We answer the following two research questions in this section:
\begin{itemize}
\vspace{-2mm}
    \item Can we detect the operational state effectively (i.e., with high accuracy)?
    \item Can we transfer the learned model across the two different types of sensors?
\end{itemize}

\subsection{DNN Model Results}

\textbf{Experimental Setup and Results}: We collected the data from 2 deployed sensors, i.e., piezoelectric and MEMS sensors mentioned earlier.  Then, two DNN models were built on these two datasets. First, a normal model for each RPM was built where we train a DNN model on around 480K samples for the RPM. We have a sampling rate of 3.2 KHz (i.e., collect 3.2K data during 1 second) and we collect 50 samples and we have 3 axes. So, total data for one experiment is $3200 \times 50 \times 3 = 480K$ data points. For testing on same RPM, the training size was 70\% of the total collected data while  the testing size was 30\%. The baseline DNN model consists of 50 neurons per layer, 2 hidden layers (with ReLU activation function for each hidden layer) and output layer with Softmax activation function. Following standard tuning of the model, we created different variants of the models to choose the best parameters (by comparing the performance of the multi-class classification problem). We built upon the Keras library \cite{gulli2017deep} which is Python-based for creating the variants of our models. In our results, we call the two models DNN-R and DNN-TL where the first refer to training DNN regularly and testing on the same sensor while the latter means transfer learning model where training was performed on one sensor and classification was performed on a different sensor (using the design of shared weights and learned representations as described in Sec.~\ref{proposed_model_transfer_learning}). Specifically, for the DNN-TL, training was done on the plentiful sensor data from the piezoelectric sensor and the prediction was done based on the MEMS sensor data. The comparison between regular DNN model and our transfer-learning DNN model on MEMS sensors in terms of the best achieved accuracy is shown in Table~\ref{tbl:comparison_results_dnn_tl}. We notice that the transfer-learning model gives a relative gain of 11.6\% over the model trained only on the lower resolution MEMS sensor data. The intuition here is that the MEMS sensor data is only 2000 samples, due to very low sampling rate (10 Hz as opposed to 3.2 kHz with the piezoelectric sensor) and thus it cannot fit a good DNN-R model.  On the other hand, we can train a DNN-TL model with sensor of different type (but still with vibration readings) with huge data and classify the failure of the sensor under test (i.e., MEMS with less data) with accuracy 71.71\%. 

\begin{table}
\caption {A comparison between regular DNN model and our transfer-learning DNN model on MEMS sensors. The transfer-learning model gives an absolute gain of 7.48\% over regulard DNN model.} 
\label{tbl:comparison_results_dnn_tl}
\centering
\small
\resizebox{0.4\textwidth}{!}{
\begin{tabular}{|c|c|c|} 
 \hline
 Model Type &  Sensor Tested & Accuracy (\%) \\ [0.5ex] 
 \hline
  DNN-R & MEMS & 64.23\% \\
  DNN-TL & MEMS &  71.71\% \\
  DNN-R & Piezoelectric &  80.01\% \\
 \hline
\end{tabular}
}
\vspace{-1mm}
\end{table}

Moreover, we show the effect of parameter-tuning on the performance of the models in Table~\ref{tbl:dnn_tuning_results}. The parameter tuning gives an absolute gain of 13.71\% over the baseline DNN-TL model. Delving into the specifics, the most effective tuning steps were feature-selection and normalization which give absolute increase of 10.66\% in the accuracy over non-normalized features and increasing number of hidden layers and batch size which gave around 3.05\% each on the performance. Note that increasing the epochs to 200 and hidden layers to more than 3 decreases the accuracy, due to over-fitting. 

\begin{table}%
\caption {The effect of parameter tuning on the accuracy of DNN-TL model. 
The parameter tuning gave an absolute gain of 13.71\% over the baseline model.} 
\label{tbl:dnn_tuning_results}
\centering
\small
\resizebox{0.7\textwidth}{!}{
\begin{tabular}{|c|c|c|c|} 
 \hline
  Tuning Factor & Accuracy & Tuning Factor & Accuracy \\ [0.5ex]
 \hline
   None & 58.00\% & Feature Selection & 64.08\% \\
  Feature-normalization & 68.66\%  &  Neurons per layer (50-80-100) & 69.41\% \\
  Number of Hidden layers (2-3) & 70.32\% &  Number of Epochs (50-100) & 70.75\% \\
  Batch Size (50-100) & \textbf{71.71\%} & & \\
 \hline
\end{tabular}
}
\vspace{-12pt}
\end{table}

\textbf{Feature Selection:} We validate one idea that the vibration data in certain axis will not carry different information in normal and failure cases. The circular movement around the center of the motor is on X and Z axes so that they have vibration values that change with motor condition while the Y-axis has smaller vibration (the direction of the shaft). Thus, we compare the result of the model when the features are the three data axes in one setup (i.e., default setup) and the proposed idea when the features are extracted only from X-axis and Z-axis data vectors.
According to the experimental setup shown in Fig.~\ref{fig:testbed1}, as the motor rotates with the disk, which is imbalanced by the mounted mass, i.e., eccentric weight, the centripetal forces become unbalanced, and this causes repeated vibrations along multiple directions. Considering the circular movement around the center of the motor, the two directions, which are x-axis and z-axis in our case, are mainly vibrated while the y-axis (the direction along the shaft) show relatively smaller vibration, which may not show a distinguishable variation in the data pattern as machine health varies. We find that this feature selection process gave us a relative increase of 10.5\%  over the baseline model with all three features. Specifically, the accuracy is 58\% using the  model trained on default features compared to 64.08\% using the model with feature selection. This kind of feature selection requires domain knowledge, specifically about the way the motor vibrates and the relative placement of the sensors.
The intuition here is that redundant data features are affecting the model's learning and therefore selecting the most discriminating features helps the neural network learning. 


\subsection{Data-augmentation Model Results} 

\textbf{Experimental Setup}: We used data-augmentation techniques (by both augmenting data from different RPMs and generating samples with interpolation within each RPM) and train DNN-R model on each sensor. For piezoelectric sensor, the data-augmentation model consists of 5M samples (480K samples collected from each rotational speed data for the available ten rotational speeds and 20K generated samples by interpolation within each RPM). For MEMS sensor, the data augmentation model consists of 15120 samples. We compare the average accuracy of the model over all RPMs under the regular model (DNN-R) and the augmented model. The absolute increase in the accuracy using the augmentation techniques over the regular model is 9.76\% for piezoelectric and 8.99\% for MEMS, respectively. The data-augmentation techniques are useful for both piezoelectric and MEMS vibration sensors. Data-augmentation is useful for transfer learning across different RPMs.

\textbf{Effect of Variation of RPMs Results:} Here, we show the details of each RPM-single model and the details of the data-augmented model. First, we train a single-RPM model and test that model on all RPMs. Then, we build a data-augmented model as explained earlier. Table~\ref{tbl:different_rpms_results} shows such comparison where the single-RPM model can't transfer the knowledge to another RPMs. An interesting note is that at the slowest RPMs (here, RPM-100 and RPM-200) the separation is harder at the boundary between failure, near-failure, and normal operational states. On the other hand, data-augmented model has such merit since it is trained on different samples from all RPMs with adding data-augmentation techniques. In details, the absolute enhancement in the average accuracy across all RPMs is 6\% while it is 13\% over the worst single-RPM model (i.e., RPM-600). 
In the data-augmented model, 70\% from each RPM's samples were selected for training that model as mentioned earlier.

\begin{table}[htbp]
\Large
\caption {Comparison on the performance of failure detection model where the trained model is using one RPM and the tested data is from another RPM. The data-augmentation model is useful for transfer the learning across different RPMs. The absolute enhancement in the average accuracy across all RPMs is 6\% while it is 13\% over the worst single-RPM model.} 
\label{tbl:different_rpms_results}
\centering
\resizebox{0.98\textwidth}{!}{
\begin{tabular}{|c|c|c|c|c|c|c|c|} 
 \hline
 Trained RPM &  RPM-100 & RPM-200 & RPM-300 & RPM-400 & RPM-500 & RPM-600 & Average (\%) \\ [0.5ex]
 \hline
  RPM-100 & 68.80\% & 66.64\% & 65.83\% & 73.61\% & 67.29\% &  42.90\% & 64.18\% \\
  RPM-200 & 63.54\% & 73.71\% & 58.11\% & 74.67\% & 67.68\% &  45.93\% & 63.94\% \\
  RPM-300 & 57.99\% & 55.00\% & 95.20\% &  66.09\% & 71.32\%    &   45.14\%	&     65.12\%  \\
  RPM-400 & 66.37\% & 69.68\% & 54.79\% &  87.38\% & 69.52\%    &  32.62\% & 63.39\%     \\
  RPM-500 & 65.37\% &  64.94\% & 80.12\% & 80.20\% & 75.61\% & 42.59\% & 68.06\%  \\
  RPM-600 & 49.16\% & 51.12\% &  63.44\% & 44.23\% &  55.02\% & 75.16\% & 56.35\%   \\
  Augmented-data model & 67.94\% & 71.31\%  & 62.61\%  & 80.06\% & 69.06\% & 65.88\% & \textbf{69.48\%}  \\
 \hline
\end{tabular}
}
\vspace{-4pt}
\end{table}

\textbf{Confusion Matrices Comparison}: Here, we show the confusion matrix which compare the performance of our DNN-R models for each operational state separately. Table~\ref{tbl:confusion_matrix_aug}(a) shows such metric using data-augmentation. The best performance is for near-failure which exceeds 96\%. This is good in practice since it gives early alarm (or warning) about the expected failure in future. Moreover, the model has good performance in normal operation which exceeds 70\%. Finally, the failure accuracy is a little lower which is 61.67\% however the confusion is with near-failure state which also gives alarm under such prediction. On the other hand, DNN-R model without data-augmentation has worse prediction in both normal and near-failure modes as shown in Table\ref{tbl:confusion_matrix_aug}(b) (normal operation detection around 60.4\% and near-failure is 93.86\%) while much better for detecting failures where the accuracy is 75.00\%. The intuition here is that detecting near-failure and normal-operation modes can be enhanced using data-augmentation techniques. On the contrary, detecting failure operational state is better without data-augmentation as failure nature can be specific for each RPM and thus creating single model for each RPM can be useful in that sense.

\begin{table}
\centering
\vspace{-8mm}
\caption{Confusion matrices for classifying operational conditions using DNN-R, where we have the following cases (a) with data-augmentation and (b) no data-augmentation.}
\label{tbl:confusion_matrix_aug}
\resizebox{0.95\textwidth}{!}{
\begin{tabular}{l|l|c|c|c||c |c |c|}%
\multicolumn{2}{c}{}&\multicolumn{2}{c}{\textit{\hspace{10mm} (a) Data Augmentation}}& \multicolumn{1}{c}{}&\multicolumn{2}{c}{\textit{\hspace{10mm} (b) No Data Augmentation}}\\
\cline{3-8}
\multicolumn{2}{c|}{}& Normal & Near-failure & Failure & Normal & Near-failure & Failure \\
\cline{2-8}
& Normal & $\textbf{70.22\%}$ & $28.40\%$ & $1.38\%$ &  $\textbf{60.38\%}$ & $18.52\%$ & $21.09\%$ \\ 
\cline{2-8}
& Near-failure & $3.82\%$ & $\textbf{96.05\%}$ & $0.13\%$ & 
$6.08\%$ & $\textbf{93.86\%}$ & $0.06\%$\\
\cline{2-8}
\cline{2-5}
& Failure & $0\%$ & $38.33\%$ & $\textbf{61.67\%}$ & $0\%$ & $25.00\%$ & $\textbf{75.00\%}$ \\
\cline{2-8}
\end{tabular}
}%
\vspace{-4mm}
\end{table}

\subsection{Relaxation of the Classification Problem}
In some applications of the sensor data, the goal can be to detect only if the data from the deployed sensor is normal or not. Thus, we relax the defect classification problem into binary classification problem to test such application. 
In this subsection, the experimental data obtained under five rotation speeds, i.e., 300, 320, 340, 360, and 380 RPMs were considered to classify between normal and not-normal states. For the deep learning model, we use neural network, which consists of two layers. 
The models’ performances are summarized in Table \ref{tbl:binary_CNN_different_rpms_results}. Compared to the original defect classification problem, the performance here is better due to the following reasons. First, the confusion is less in binary classification problem since we have only two classes. Second, the variation in the range between RPMS is less in this experiment. 
 \begin{table}[htbp]
 \caption {Comparison on the performance of binary classifier detection model where the trained model is using one RPM and the tested data is from another RPM.  The average accuracy is higher, compared to the three classes defect classification models.} 
\label{tbl:binary_CNN_different_rpms_results}
\centering
\resizebox{0.78\textwidth}{!}{
\begin{tabular}{|c|c|c|c|c|c|c|} 
 \hline
 Trained RPM &  RPM-300 & RPM-320 & RPM-340 & RPM-360 & RPM-380 &  Average (\%) \\ [0.5ex]
 \hline
  RPM-300 & \textbf{100\%} & 65.17\% & 58.17\% & 51.50\% &  50.63\% & 65.09\% \\
  RPM-320 & 99.75\% & \textbf{100\%} & 97.58\% & 78.63\% & 68.17\% &  88.82\% \\
  RPM-340 & 96.60\% & 99.27\% & \textbf{100\%} &  97.33\% & 82.43\%    &  95.12\%  \\
  RPM-360 & 96.60\% & 99.27\% & 97.33\% &  \textbf{99.67\%} & 84.43\% & \textbf{95.46\%} \\
  RPM-380 & 61.05\% &  87.75\% & 96.93\% & 99.83\% &  \textbf{99.77\%} & 89.07\%  \\
 \hline
\end{tabular}
}
\vspace{-6mm}
\end{table}

\section{Related Work}\label{sec:lit-review}

\noindent {\bf Failure detection Models}:
There have been several works to study a failure detection in manufacturing processes using single or multi-sensor data \cite{lee2019mfg2,teng1996failure,lee2019mfg3}. Specifically, the recent work \cite{lee2019mfg2}, in which the kernel principal component analysis based anomaly detection system was proposed to detect a cutting tool failure in a machining process. In the study, multi-sensor signals were used to estimate the condition of a cutting tool, but a transfer learning between different sensor types was not considered. Also, in another recent study \cite{lee2019mfg3}, the fault detection monitoring system was proposed to detect   various failures in a DC motor  such as a gear defect, misalignment, and looseness. In the study, a single sensor, i.e., accelerometer, was used to obtain machine condition data, and several convolutional neural network architectures were used to detect the targeted failures. However, different rotational speeds and sensors were not considered. Thus, these techniques must be applied again for each new sensor type. On the other hand, we consider the transfer learning between different sensor types. We also compare traditional and ML-based models for our anomaly detection task.

\noindent {\bf Learning Transfer}: 
Transfer learning has been proposed to 
extract knowledge from one or more source tasks and apply the knowledge to a target task \cite{xiang2010bridging,qiu2016survey,torrey2010transfer} with the advantage of intelligently applying knowledge learned previously to
solve new problems faster. In the literature, transfer learning techniques have been applied successfully in many real-world data processing applications, such as cross-domain text classification, constructing informative priors, and large-scale document classification \cite{ling2008spectral,chen2015net2net}. However, these works did not tackle the transfer the learning across different instances of sensors that we consider here and did not consider the smart manufacturing domain. In smart manufacturing systems, the existing works only considered calibration of sensors using neural network regression models~\cite{9181609} and multi-fault Bearing classification~\cite{8868093}. However, these works did not tackle the transfer of the learning across different instances of sensors that we consider here.

\section{Discussion and Limitations}

\textbf{Ethical concerns:} We do not see significant risks of security threats or human rights violations in
our work or its potential applications. However, we do foresee that our work contributes to the field
of smart manufacturing and anomaly detection fields overall. These efforts might eventually automate the detection process, leading to changes in the workforce structure. Hence, there is a general concern that automation may significantly reduce the demand for manufacturing human workers, and the industries would need to act proactively to avoid the social impact of such changes.

\textbf{Transfer Learning under Different Features:}
In our transfer learning task, the sensor types I and II should be measuring the same physical quantity but can be from different manufacturers and with different characteristics. Another interesting question would be what happens if the two sensor types have overlapping but not identical features in the data that they generate? This requires more complex models which can do feature transformations, using possibly domain knowledge, and we leave such investigation for future work.

\textbf{Reproducibility:}
We have publicly released our source codes and benchmark data to enable others  reproduce our work. We are publicly releasing, with this submission, our smart manufacturing database corpus of 4 datasets. This resource will encourage the community to standardize efforts at benchmarking anomaly detection in this important domain. We  encourage the community to expand this resource by contributing their new datasets and models. The website with our database and source codes is:  \url{https://drive.google.com/drive/u/2/folders/1QX3chnSTKO3PsEhi5kBdf9WwMBmOriJ8}. 
The details of each dataset and the different categories of models are in Section~\ref{sec:anomaly_detect_manuf}. The hyper-parameter selections and the libraries used are presented in Appendix E.

\section{Conclusion}\label{sec:conclusion}

This paper explored several interesting challenges to an important application area, smart manufacturing. We studied \emph{anomaly detection} and \emph{failure classification} for the predictive maintenance problem of smart manufacturing. We designed a temporal anomaly detection technique and an efficient defect-type classification technique for such application domain. We compared the traditional and ML-based models for anomaly detection. We observed that ML-based models lead to better anomaly detection prediction. We tested our findings on four real-world data-sets. We then proposed a transfer learning model for classifying failure on sensors with lower sampling rate (MEMs) using learning from sensors with huge data (piezoelectric) where the model can detect anomalies across operating regimes. Our findings indicate that the transfer learning model can considerably increase the accuracy of failure detection. We also studied the effects of several tuning parameters to enhance the failure classification. We release our database
corpus and codes for the community to build on it with new datasets and models.
Future avenues of research include  leveraging the data from multiple sensors and detecting the device health by merging information from multiple, potentially different, sensors. 

\section{Acknowledgment}

This work is supported by the Wabash Heartland Innovation Netowrk (WHIN). The opinions expressed in this publication are those of the authors. They do not purport to reflect the opinions or views of sponsor. We thank Nithin Raghunathan and Ali Shakouri for their valuable feedback and suggestions on the work.

\bibliography{sample}
\bibliographystyle{plain}

\normalfont

\appendix 
\section*{Appendix}
The supplementary material of our work is organized as follows.

\section{Organization of the Appendix}
\begin{itemize}
    \item[] (B) Reproducibility and URL for Datasets and Codes
    \item[] (C) Author Statement, Hosting,  and Dataset License
    \item[] (D) Explaining Datasets: Highlights, and How Can it be Read
    \item[] (E) Benchmarks: Models,  Hyper-parameter Selection, and Code Details
    \item[] (F) Extended Evaluation
    \item[] (G) Main Additions compared to Preliminary Version 
    \end{itemize}

\section{Reproducibility and URL for Datasets and Codes}

We have publicly released our source codes and benchmark data to enable others  reproduce our work. In particular,  we are publicly releasing, with this submission, our smart manufacturing database corpus of 4 datasets. This resource will encourage the community to standardize efforts at benchmarking anomaly detection in this important domain. We also encourage the community to expand this resource by contributing their new datasets and models. The website with our database and source codes is:  \url{https://drive.google.com/drive/u/2/folders/1QX3chnSTKO3PsEhi5kBdf9WwMBmOriJ8}. 
The details of each dataset and the different categories of models are in Appendix~\ref{app:datasets_details} and Appendix~\ref{app:hyper_param_model_space}, respectively. We provide the datasheet for the datasets in the supplementary material. The hyper-parameter selections and the libraries used for each model are presented in Appendix~\ref{app:hyper_param_model_space}.

\section{Author Statement, Hosting,  and Dataset License} 

We bear all responsibility in case of violation of rights, etc., and below we confirm the consent from a pharmaceutical packaging manufacturer company and show the data license. We also show our hosting of the data (and code) and maintenance.

\subsection{Consent from Pharmaceutical Packaging}\label{app:consent}

For our testbeds, we have created the testbeds for MEMs, Piezoelectric, and Process data from real-data sensors. On the other hand, we get approval from the Pharmaceutical Packaging company which provided us with their vibration data which we used as one dataset for our anomaly detection task. The approval was taking by sending us electronic email confirming using and publishing this dataset under such name. We clarified such a thing in the datasheet section above.

\subsection{Hosting and Maintenance}
To ensure accessibility and future maintenance of the datasets and the codes. We created a Github repository with the following URL\footnote{Github Code and Dataset Link:\\ \url{https://github.com/submission-2022/Smart-Manufacturing-Testbed-for-Anomaly-Detection.git}}. It contains all the codes, datasets, and most of the instructions. We chose a creative common Zero 1.0 Universal license for the repository. We will use that github repository to update the code and datasets and enhance them.

\section{Explaining Datasets: Highlights, and How Can it be Read}\label{app:datasets_details}

\subsection{MEMs and  Piezoelectric datasets:}

\textbf{Highlights of the datasets:}
To build these datasets, an experiment was conducted in the motor testbed to collect machine condition data for different health conditions. During the experiment, the acceleration signals were collected from both piezoelectric and MEMS sensors at the same time with the sampling rate of 3.2 kHz and 10 Hz, respectively, for X, Y, and Z axes. Different levels of machine health condition was induced by mounting a mass on the balancing disk, thus different levels of mechanical imbalance are used to trigger failures. Failure conditions were classified as one of three possible states - normal, near-failure, and failure. In this experiment, three levels of mechanical imbalance (i.e., normal, near-failure, failure) were considered acceleration data were collected at the ten rotational speeds (100, 200, 300, 320, 340, 360, 380, 400, 500, and 600 RPM) for each condition. While the motor is running, 50 samples were collected at a 10 second interval, for each of the ten rotational speeds.

\textbf{Reading the dataset:}
Both Piezoelectric and MEMs databases are in CSV format. For Anomaly detection, we have a single RPM (CSV file) while for transfer learning we have several rpms for each RPM (where all CSV files are compressed in .zip format). The CSV file for Piezoelectric has many more samples (due to higher sampling rate). For each CSV file, each data instance (row) contains the following columns: X, Y, Z  where each one has the corresponding vibration sensor reading.

\subsection{Process Data}\label{sec:process_data}
\textbf{Highlights of the dataset:}
\begin{itemize}
    \item \textit{Start date:} 7/27/2021
    \item \textit{End date:} 5/1/2022 
    \item \textit{Measurement Columns:} Air Pressure 1, Air Pressure 2, Chiller 1 Supply Tmp, Chiller 2 Supply Tmp, Outside Air Temp, Outside 
  Humidity, and Outside Dewpoint.
  \item \textit{Measurement interval:} 5 mins (1 data point per 5 min)
  \item \textit{Description:} We were also able to collect process related data that can potentially indicate the condition of machine operation. We call it process data and the process data has been collected with the Statistical Process Control (SPC) system. The measurement started from Aug. 2021 until May 2022.
\end{itemize}

\begin{figure}[ht]
\centering
  \includegraphics[width=0.85\linewidth]{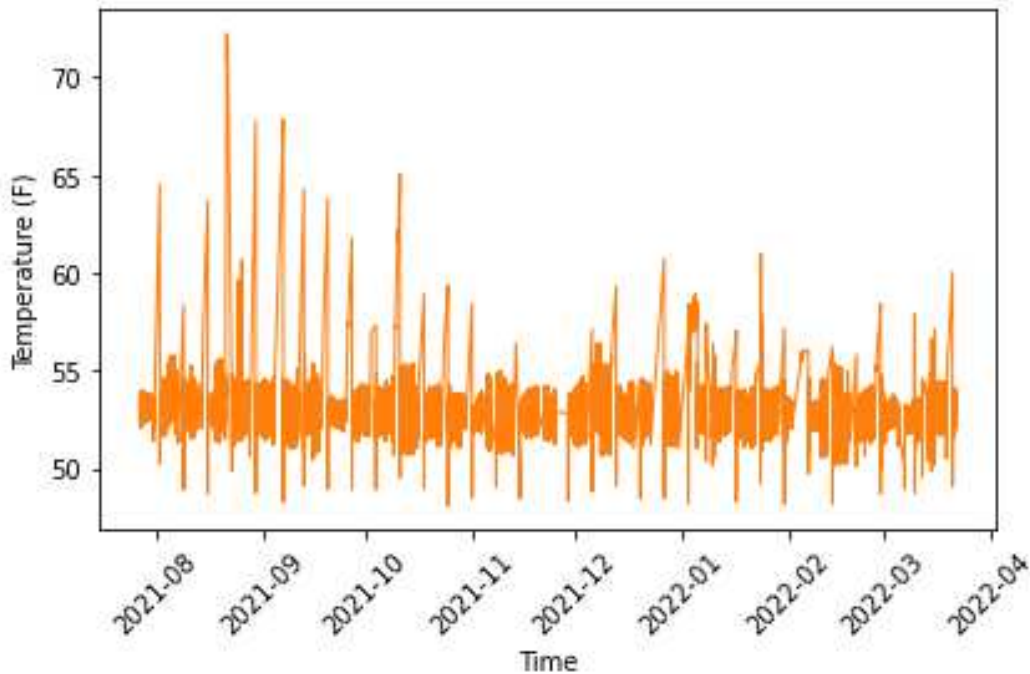}
  \caption{Chiller supply temperature in the process data}
  \label{fig:processdata}
\end{figure}

\textbf{Reading the dataset:} The Process data is in CSV format. The CSV file has around 49K instances where each data instance (row) contains the following columns: Timestamp, Air Pressure 1, Air Pressure 2, Chiller 1 Supply Tmp, Chiller 2 Supply Tmp, Outside Air Temp, Outside Humidity, and Outside Dewpoint. We applied our anomaly detection techniques on chiller supply temperature. Figure~\ref{fig:processdata} shows one of the process data, chiller supply temperature.

\textbf{Abnormal Dates:}
We observed abnormal operations of the machine which occurred on February 1st 2022 and March 8th 2022.

\subsection{Pharmaceutical Packaging}

\textbf{Highlights of the dataset:}
\begin{itemize}
\item \textit{Start date:} 11/13/2021, some data loss between December and January.
\item \textit{End date:} May 2022
\item  \textit{Measurement location:} Air Compressor, Chiller 1, Chiller 2, and Jomar moulding machine
\item \textit{Sample rate for each axis:} 3.2 kHz
\item \textit{Measurement interval:} 30 mins
\item \textit{Measurement duration:} 1 seconds
\end{itemize}

\textbf{Reading the dataset}
The dataset have several months where each month is represented by a ".txt" file that indicates vibration data for each period. In particular, the vibration data for one measurement is written into 5 lines as follows:
\begin{itemize}
    \item 1st line: date and time that the measurement started
    \item 2nd line: x-axis vibration data (3200 data points)
    \item 3rd line: y-axis vibration data (3200 data points)
    \item 4th line: z-axis vibration data (3200 data points)
    \item 5th line: time difference between each data point
\end{itemize}

\section{Benchmarks: Models,  Hyper-parameter Selection, and Code Details}\label{app:hyper_param_model_space}

\subsection{Models and Hyper-parameter Selection}
{
We now provide details on the models used to study the anomaly detection problem in our work. We explain
the time-series forecasting algorithm and the hyperparameters used and the libraries used for each forecasting model. This can help reproducing our results for the future related works.
}

{
\noindent \textbf{DeepAR~\cite{salinas2020deepar}:} DeepAR experiments are using the model implementation provided by GluonTS version 1.7. We did grid search on different values of  \textit{number of cells} and \textit{the number of RNN layers}  hyperparameters of DeepAR since the defaults provided in GluonTS would often lead to apparently suboptimal performance on many of the datasets. The best values for our parameters are number of cells equals 30 and number of layers equals 3.  All other parameters are defaults
of gluonts.model.deepar.DeepAREstimator.
}

{
\noindent \textbf{Deep Factors~\cite{wang2019deep}:} Deep Factors experiments are using the model implementation provided by GluonTS version 1.7. We did grid search over \textit{the number of units per hidden layer} for the global RNN model  and \textit{the number of global factors} hyperparameters of Deep Factors. The best values for our parameters are 30 (for the number of units per hidden layer) and 10 (for the number of global factors).  All other parameters are defaults
of gluonts.model.deep\_factor.DeepFactorEstimator.
}

{
\noindent \textbf{Seasonal Naive~\cite{montero2020fforma}:} Seasonal Naive experiments are using the model implementation provided by GluonTS version 1.7. We did grid search over \textit{the length of seasonality pattern}, since it is different unknown for each dataset. The best parameter was either 1 or 10 for all datasets.  All other parameters are defaults
of gluonts.model.seasonal\_naive.SeasonalNaivePredictor.
}

{
\noindent \textbf{Auto Regression~\cite{lewis1985prediction}:} Auto Regression experiments are using the model implementation provided by statsmodels python library version 0.12.2. We did grid search over the \textit{loss covariance type} and \textit{the trend} hyperparameter of Vector Auto Regression. The best parameters are `HC0' (for loss covariance type) and `t' (for trend hyper-parameter).  All other parameters are defaults
of statsmodels.tsa.var\_model.
}

{
\noindent \textbf{Random Forest~\cite{liaw2002classification}:} Random Forest models' experiments are using the model implementation provided by sklearn python library version 0.24.2. We did grid search over \textit{the number of estimators (trees)} and \textit{the max\_depth (i.e., the longest path between the root node and the leaf node in a tree)} hyperparameter of Random Forest. The best  parameters are 500 (for the number of estimators) and 10 (for the  max\_depth). All other parameters are defaults
of sklearn.ensemble.RandomForestRegressor.
}

{
\noindent \textbf{ARIMA~\cite{contreras2003arima}:} ARIMA model experiments are using the model implementation provided by the statsmodels python library version 0.12.2. A typical ARIMA model can be represented as a function ARIMA(p,d,q) where $p$ is the the number of lag observations included in the regression model, $d$ is the number of times that the raw observations are differenced, and $q$ is the size of the moving average window. Then, we use this trained ARIMA model to detect the anomaly in the sensor’s test (future) readings. In practice, $p = 0$
or $q = 0$ as they may cancel each other. The best parameters in our experiments were $q = 0, d = 1$ and $p = 10$ after tuning trials. All other parameters are defaults
of statsmodels.tsa.arima.model. The reason for our choice of ARIMA is that if the data has a moving average linear relation (which we estimated the data does have), ARIMA  would be better to model such data. Moreover, ARIMA is a simple and computationally efficient model.
}

{
\noindent \textbf{Simple RNN~\cite{tokgoz2018rnn}:} Simple Recurrent Neural Network (RNN) models experiments are using the model implementation provided by keras python library version 2.9.0. We did grid search over several parameters. The best parameters are 100 neurons per layer with `Relu' activation function. We have two hidden layers with also `Relu' activation. We used batch size of 10. All other parameters are defaults
of keras.layers.SimpleRNN.
}

{
\noindent \textbf{LSTM~\cite{graves2005framewise}:} Long-short Term Memory (LSTM) models experiments are using the model implementation provided by  keras python library version 2.9.0. LSTM better models data that has non-linear relationships, which is suitable for manufacturing sensors' readings and, thus, LSTM can be a more expressive model for our anomaly detection task. We did grid search over several parameters. The best parameters are 100 neurons per layer with `Relu' activation function. We have two hidden layers with also `Relu' activation. We used batch size of 10. We used 4 LSTM blocks and one dense LSTM layer with 10 units and the training algorithm used is Stochastic Gradient Descent (SGD). All other parameters are defaults of keras.layers.LSTM.
}

{
\noindent \textbf{AutoEncoder~\cite{chollet2016building}:} AutoEncoder models' experiments are using the model implementation provided by keras python library version 2.9.0. We did grid search over several parameters. The best parameters we have are: using 3 convolutional layers where each layer has 32 filters, 2 strides, kernel size of 7, and `Relu' activation. The dropout rate is 0.2.  We used batch size of 10. All other parameters are defaults of keras.layers.Sequential and keras.layers.Conv1D.
}

{
\noindent \textbf{DNN~\cite{sen2019think}:} Deep Neural Network (DNN) models experiments are using the model implementation provided by keras python library version 2.9.0. We did grid search over \textit{the number of layers}, \textit{batch size}, and \textit{number of neurons per layer}. The best parameters are 50 neurons per layer with `Relu' activation function. We have three hidden layers with also `Relu' activation. We used batch size of 10. All other parameters are defaults of keras.layers.Dense.
}

\subsection{Code Details and Prerequisites}
We share our codes along with database corpus. In particular, the code's link is \url{https://drive.google.com/drive/u/2/folders/14eY8tsr-PALnifSQpEqlXgr_RVstaicd}. The codes folder is divided into two sub-folders:
(1) Anomaly Detection Codes and (2) Transfer Learning Codes, which are detailed below.

\subsubsection{Anomaly Detection Code}
Under Anomaly detection folder, we have the following source codes:
\begin{itemize}
    \item \textbf{Arima.py:} Training and testing ARIMA forecasting model 
    \item \textbf{LSTM.py:} Training and testing LSTM forecasting model 
    \item \textbf{AutoEncoder.py:} Training and testing AutoEncoder forecasting model 
    \item \textbf{DNN.py:} Training and testing DNN forecasting model 
    \item \textbf{RNN.py:} Training and testing RNN forecasting model 
    \item \textbf{GluonTModels.py:} That file contains the rest of the forecasting models along with the required functions and hyper-parameters.
    \item \textbf{Autoencoder\_classifier.py}: Training and Testing a classifier-based model for  anomaly detection (See Appendix F.2).
\end{itemize}

In particular, "GluonTModels.py" contains the codes to train and examine performance for the following models:
\begin{itemize}
    \item \textbf{DeepAR}
    \item \textbf{DeepFactors}
    \item \textbf{Seasonal Naive}
    \item  \textbf{Random Forest}
    \item  \textbf{Auto-Regression}
\end{itemize}


\subsubsection{Transfer Learning Code}
Under Transfer learning folder, we have the following source codes:
\begin{itemize}
    \item \textbf{Transfer\_Learning\_Pre\_processing.py:} This code prepares the CSV files used for training defect type classifiers and transfer learning. 
    \item \textbf{Transfer\_Learning\_Train.py:} This code trains and tests the defect type classifier, including feature encoding, defect classifier building, testing, and performance reporting. 
\end{itemize}

\subsubsection{Running the Codes}
To run any code, we need just to run the command “python code\_name.py”, where "code\_name" is the required anomaly detection or transfer learning model. The user would need to change the datafile name inside the code to the dataset of choice.

\subsubsection{Prerequisites (Libraries and Modules)}
Our codes have the following libraries that need to be installed (which can be installed using apt-get install or conda): 
\begin{itemize}
    \item numpy, scipy, pandas, and sklearn, re, random, and csv
    \item GluonTS, keras, statsmodels, matplotlib, and simplejson
\end{itemize}

\section{Extended Evaluation}\label{app:extended_evaluation}

\subsection{Anomaly Detection using Autoencoder Classification}\label{sec:autoencoder_anomaly_classification}

For some manufacturing sensors (such as process data in our paper), the classification is changed from normal, failure, and near-failure (warning) to running, stopped, and abnormal due to working hours for such manufacturing facilities. Thus, in this section, we will use autoencoder classification for such operation state on process data (described in Appendix~\ref{sec:process_data}).

\begin{figure}[ht]
 \vspace{-3mm}
\centering
  \includegraphics[width=0.85\linewidth]{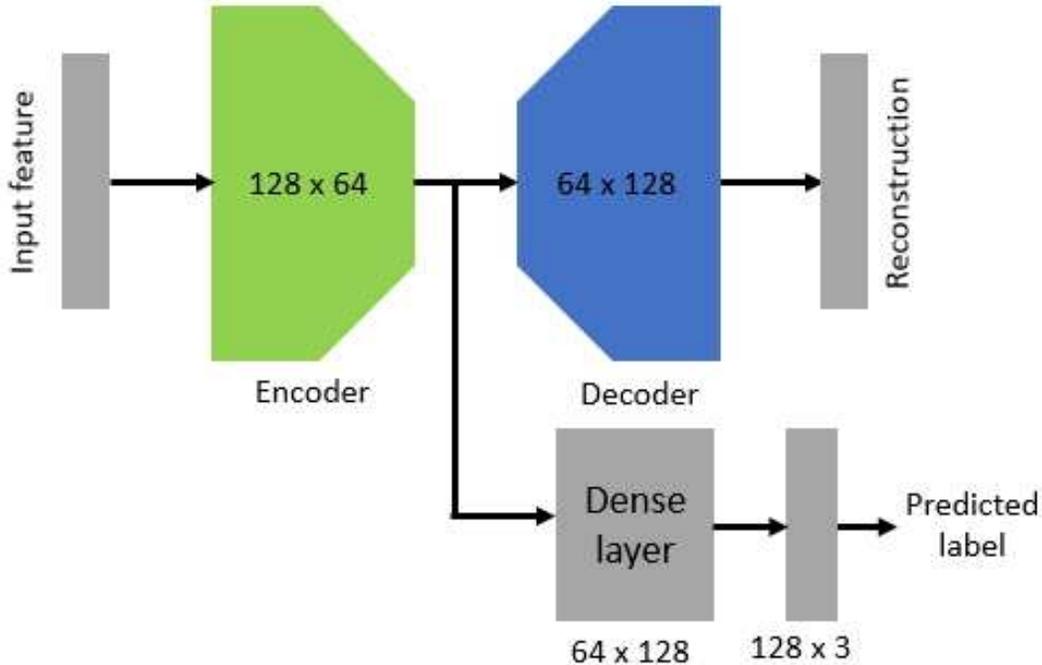}
  \caption{Proposed autoencoder model with an additional classification layer}
  \label{fig:autoencoder}
  \vspace{-3mm}
\end{figure}

\textbf{Autoencoder Classifier}: An autoencoder is used for the classification of machine operation states (running, stopped, abnormal). We employed a simple autoencoder of which the encoder and the decoder are consisting of a single hidden layer and an output layer with an additional classification layer. As shown in Figure~\ref{fig:autoencoder}, the encoder consists of two linear layers (128, 64) and the decoder consists of another two linear layers (64, 128). The classification layer also has two linear layers (128, 3) that outputs the predicted label. Here, the output of the classification layer is one of the three different operation conditions. The autoencoder can have two different losses (reconstruction loss and classification loss) during the training. We developed a loss function to minimize a weighted sum of the two losses. The best number of layers in the autoencoder has been determined to be 2. We also have observed that the depth of the autoencoder does not improve the prediction accuracy. 

\subsection{Experimental Setup and Results
}
An autoencoder was developed to perform experiments for the anomaly detection on tri-axial vibration data and process data. The vibration data and process data were not lined up as they are measured from different sources. We performed label imputation to line up the timestamp of both vibration and process data. Here, median values are utilized. Then, we extracted time domain features from the vibration data to construct the autoencoder's input. The main features used for that task were: mean, standard deviation, root mean square, peak,and crest factor in the feature extraction. The final input consists of the extracted time domain features and process data which indicates the operation of the manufacturing equipment. Then, we performed labeling task based on the machine operation information as follows. As the pharmaceutical company operates non-stop from Sunday 11 pm to Friday 7 pm and shuts down from Friday 7 pm to Sunday 11 pm.  When the chiller is failed, supply temperature can vary with the ambient temperature and goes up to 65 degrees. When the machine is off, the data is labeled as `0' whereas the label is determined as `1' when the machine is on. When there are abnormal operation of the machine, the data is labeled as `2'. During the data collection, we observed two abnormal operations of the machine which are occurred on February 1st 2022 and March 8th 2022. When the abnormal operations were detected, maintenance was performed to lubricate the machine. Data collection was being conducted when the machine was under the maintenance service. We aim to detect such abnormal operations using the vibration and process data with our proposed model. The proposed model achieved 84\% of test accuracy. The limited number of abnormal labels affected the accuracy.

\section{Main Additions compared to Preliminary Version}

We have submitted an earlier version of this work~\cite{abdallah2021anomaly}. Below we list the main additions and enhancement over that version.

\begin{itemize}
    \item Added the Precision, Recall, and F-1 score to the metrics being used in our evaluation.

    \item Provided details of feature selection (including which features are most useful for the anomaly detection and defect type classification) and model tuning (including hyper-parameter selection). 
    
    \item Added to the manufacturing datasets and related experiments by using two more recent datasets (Process data and Pharmaceutical packaging manufacturer company), with identified real failures.
    
    \item Added more recent related works and missing surveys suggested by the reviewers.
    
    \item Added limitations and Discussion section of the current work. Also, added in the datasheet the prospective usages for our datasets.
    
    \item Added eight other forecasting models from both traditional and ML-based categories to benchmark their performance on the datasets.
    
    \item Shared the datasets and the codes with URL along with detailed explanations and prospective hosting plan (with Github repository).
\end{itemize}

\end{document}